\documentclass{CVIS}
\usepackage{amsmath,bm}
\usepackage{graphicx}
\usepackage{dcolumn}
\usepackage{soul}
\usepackage{caption}
\usepackage[numbers,sort&compress,sectionbib]{natbib}
\usepackage[pagebackref=flase,breaklinks=true,letterpaper=true,colorlinks=true,linkcolor = blue, urlcolor = black,citecolor=black,bookmarks=true]{hyperref}
\usepackage{url}
\usepackage{mleftright}
\usepackage{multirow}
\usepackage{subfig}

\usepackage{titlesec}

\title{Where Does Trust Break Down? A Quantitative Trust Analysis of Deep Neural Networks via Trust Matrix and Conditional Trust Densities
}

\begin{document}
\sloppy
\author{
\begin{tabularx}{\textwidth}{X X}
Andrew Hryniowski$^{*,1,2,3}$ & $^{1}$ Vision and Image Processing Research Group\\
Xiao Yu Wang$^{1,3}$ &  $^{2}$ Waterloo Artificial Intelligence Institute, University of Waterloo\\
Alexander Wong$^{1,2,3}$ & $^{3}$ DarwinAI Corp.\\
 {*Email: andrew@darwinai.ca} & \\
\end{tabularx}
}

\maketitle
\begin{abstract}
The advances and successes in deep learning in recent years have led to considerable efforts and investments into its widespread ubiquitous adoption for a wide variety of applications, ranging from personal assistants and intelligent navigation to search and product recommendation in e-commerce.  With this tremendous rise in deep learning adoption comes questions about the trustworthiness of the deep neural networks that power these applications.  Motivated to answer such questions, there has been a very recent interest in trust quantification.  In this work, we introduce the concept of \textbf{trust matrix}, a novel trust quantification strategy that leverages the recently introduced question-answer trust metric by Wong et al. to provide deeper, more detailed insights into where trust breaks down for a given deep neural network given a set of questions.  More specifically, a trust matrix defines the expected question-answer trust for a given actor-oracle answer scenario, allowing one to quickly spot areas of low trust that needs to be addressed to improve the trustworthiness of a deep neural network.  The proposed trust matrix is simple to calculate, humanly interpretable, and to the best of the authors' knowledge is the first to study trust at the actor-oracle answer level.  We further extend the concept of trust densities with the notion of \textbf{conditional trust densities}.  We experimentally leverage trust matrices to study several well-known deep neural network architectures for image recognition, and further study the trust density and conditional trust densities for an interesting actor-oracle answer scenario. The results illustrate that trust matrices, along with conditional trust densities, can be useful tools in addition to the existing suite of trust quantification metrics for guiding practitioners and regulators in creating and certifying deep learning solutions for trusted operation.	
\end{abstract}

\section{Introduction}
The advances in deep learning~\cite{lecun2015deep} in recent years have led to considerable successes across various fields from language~\cite{vaswani2017attention,devlin2018bert,brown2020language} to vision~\cite{krizhevsky2012imagenet,ResNet50,chen2017rethinking} to medicine~\cite{wallach2015atomnet,wang2020covidnet}.  As a result, there has been considerable efforts and investments into the widespread ubiquitous adoption of deep learning for a wide variety of applications, ranging from personal assistants and autonomous driving~\cite{huang2020autonomous} to stock trend prediction~\cite{hu2017listening}.  With this tremendous rise in deep learning adoption comes questions about the trustworthiness of the deep neural networks that power these applications.  Trust in deep learning is particularly important in mission-critical areas such as healthcare, finance, and security, where decisions have considerable socioeconomic implications and the use of automated decision-making tools is regulated.

Motivated to answer such questions, there has been a very recent interest on the are of trust quantification, where the goal is to quantify the level of trustworthiness of a deep neural network in the decisions that it makes.  Given that trust quantification is a very new area of research, it is not very well explored and as such there is a limited number of early studies in research literature.  A majority of studies explore answer-level trust quantification, where the focus is to quantify the trustworthiness of an individual answer given by a deep neural network.  Answer-level trust quantification methods include: 1) uncertainty estimation strategies~\cite{titensky2018uncertainty,geifman2018biasreduced,NIPS2017_7141,gal2015dropout} which quantify the uncertainty of an individual prediction are based on statistical distributions over possible predictions, 2) agreement-based strategies ~\cite{jiang2018trust} which quantify the agreement between the classifier and a modified nearest-neighbor classifier on an individual answer, and 3) subjective logic inspired strategies~\cite{deeptrust} that construct probabilistic logic descriptions of deep neural networks and producing trust probabilities around a neural network's prediction.

More recently, there has been a growing interest in network-level trust quantification, where the goal is to quantify the overall trustworthiness of deep neural networks.  More specifically, in a study by Wong et al.~\cite{alex2020really}, a suite of trust quantification metrics were introduced to quantify and get a deeper understanding of the overall trustworthiness of deep neural networks at different levels of granularity. From most fine grain to most coarse grain these metrics include
\begin{enumerate}
    \item \textbf{Question-Answer Trust}: The function used to measure a model's trustworthiness for a single question-answer scenario in a human interpretable manner. The function is defined by $Q_z(x,y)$ which takes a question $x$, a model $M$'s answer $y$ to the question $x$, and an oracle $O$'s answer $z$ to question $x$.
    \item \textbf{Trust Density}: The trust density is the distribution of question-answer trust for a given scenario $z$. Trust density provides the ability to see where trust breaks down, and where trust holds up allowing for deeper investigation in to model-data relationships.
    \item \textbf{Trust Spectrum}: The expected trust density $T_M(z) = E[Q_z(x,y)]$ (or  trust spectrum coefficient) for a given scenario $z$ provides a method for measuring the overall trust one can place on a model $M$ for that scenario $z$. Trust Spectrum is the set of coefficients across all scenarios $z$.
    \item \textbf{NetTrustScore}: The NetTrustScore metric is the integral of a model-dataset trust spectrum. It provides a single metric indicating how trustworthy a model is overall.

\end{enumerate}

While the suite of trust quantification metrics proposed in~\cite{alex2020really} covers a broad spectrum of granularity, one particular aspect that was not investigated was trust quantification at the actor-oracle answer scenario (e.g., $(y,z)$).  We hypothesize that this particular level of granularity can complement the aforementioned suite of trust quantification metrics by providing valuable detailed insights into where trust breaks down for a given deep neural network in the context of specific individual decision scenarios, particularly for decision scenarios where actor answers misaligned with oracle answers.  Motivated by this, in this work we introduce the concept of \textbf{trust matrix}, a trust quantification mechanism designed to allow one to quickly spot areas of low trust that needs to be addressed to improve the trustworthiness of a deep neural network.   We further extend the concept of trust densities proposed in~\cite{alex2020really} with the notion of \textbf{conditional trust densities}.

The paper is organized as follows.  In Section~\ref{method}, we will briefly review the concept of question-answer trust for measuring the trustworthiness of individual actor answers with respect to oracle answers, and then present in detail the mathematical formulation of the proposed concepts of trust matrix and conditional trust densities.  In Section~\ref{results}, we will experimentally leverage trust matrices to study several well-known deep neural network architectures designed for the purpose of image recognition to investigate where trust may break down for these networks, as well as study the trust density and conditional trust densities for an interesting actor-oracle answer scenario identified using one of the trust matrices.

\section{Method}
\label{method}

Let us first review the concept of question-answer trust for measuring the trustworthiness of individual actor answers with respect to oracle answers, and the mathematical formulation behind it.  In~\cite{alex2020really}, two key logical assumptions were made about the trustworthiness of an actor from the perspective of an oracle: 1) The more confident an actor is about their wrong answer, the less trust one has in the actor, and 2) The less confident an actor is about their right answer, the less trust one has in the actor.  These assumptions align with recent social psychology studies~\cite{Tenney,Tenney07,Tenney08} that showed actors who were overconfident (i.e., high confidence coupled with known poor performance) were evaluated less positively and actors who are less confident were evaluated less positively as well.
\begin{figure*}[t]
\centering
	\includegraphics[width = 0.8\linewidth]{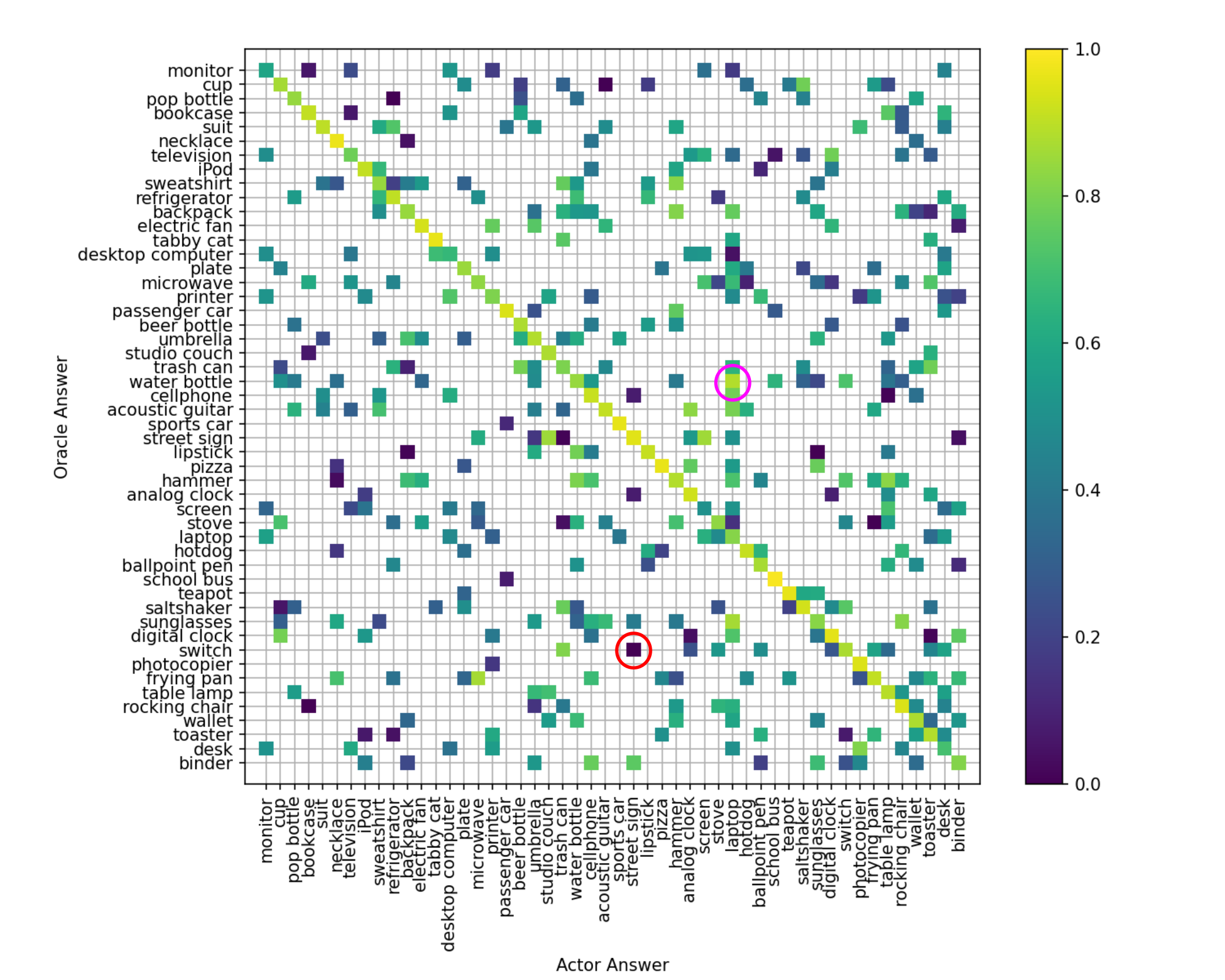}
	\caption{The trust matrix of ResNet-50 on a subset of the ImageNet~\cite{ImageNet} dataset.  Red circle marks an overconfident actor-oracle answer scenario, while pink circle marks a scenario of strong trust despite being an incorrect answer scenario.}	
	\label{fig:matrix_resnet50}
\vspace{-0.21 cm}
\end{figure*}
Based on these two assumptions, the relationship for an question-answer pair $(x,y)$ with respect to model $M$ can be defined as
\begin{equation}
y = M(x)
\end{equation}
\noindent where $x \in X$ represents the question, $X$ represents the space of all possible questions, $y \in Z$ represents the actor answer, $z \in Z$ represents the oracle answer, and $Z$ is the space of all possible answers.  We now define $R_{y \neq z|M}$ as the space of all questions $x$ where the answer $y$ by model $M$ does not match the oracle answer $z$ (i.e., incorrect answers), and $R_{y = z|M}$ as the space of all questions $x$ where the answer $y$ by model $M$ matches the oracle answer $z$ (i.e., correct answers).  Furthermore, we define $C(y|x)$ as the confidence of $M$ in an answer $y$ to question $x$.

Given the above definitions, the question-answer trust $Q_{z}(x,y)$ for a given question-answer pair $(x,y)$ can be expressed as
\begin{equation}
    Q_{z}(x,y) =
    \begin{cases}
      C(y|x)^\alpha & \text{if $x \in R_{y = z|M}$ } \\
      (1 - C(y|x))^\beta & \text{if $x \in R_{y \neq z|M}$ } \\
    \end{cases}
    \label{questiontrust}
\end{equation}
\noindent where $\alpha$ and $\beta$ denote reward and penalty relaxation coefficients.

The dynamic range of $Q_{z}(x,y)$ is [0,1], with 0 being the lowest level of trust in the answer and 1 being the highest level of trust.

\subsection{Trust Matrix}

Question-answer trust and trust densities provide a useful mechanisms for understanding if a model is trustworthy on a per-sample basis, and how the trustworithness can vary within single scenario, respectively. What these metrics fail to provide is an understanding of how a model is relating specific scenarios to one another, and how trustworthiness varies between scenario predictions for a specific scenario. Confusion matrices meet the former condition but do not help with understanding trustworthiness variation between scenario predictions.

To help understand both inter-scenario relationships on a trustworthiness level we propose the concept of the \textbf{trust matrix}, a novel trust quantification strategy for studying in detail the overall trust of a deep neural network at the actor-oracle answer level.  Let $R_{y,z}$ denote the space of questions contained within the unique actor-oracle answer tuple $(y,z)$ for a given oracle answer $z$ and actor answer $y$.

A trust matrix $Q$ is a matrix of expected question-answer trusts for every possible actor-oracle answer scenario $(y,z)$.  As such, a single element in the trust matrix (denoted here as $Q_z(y)$) for the unique actor-oracle answer tuple $(y,z)$ is defined as
\begin{equation}
    Q_z(y) = E_{x \sim P(X)}[Q_z(x,y)]
\end{equation}
\noindent where $P(X)$ denotes the probability distribution of sampling a given question $x$ from the set of questions $X$.

It can be seen that the proposed trust matrix is simple to calculate and humanly interpretable, making it well suited for providing detailed insights into where trust breaks down at the actor-oracle answer level.  In this study, we set $\alpha=1$ and $\beta=1$ to penalize undeserved overconfidence and reward well-placed confidence equally.

\subsection{Conditional Trust Densities}
Extending upon the concept of trust densities first introduced in~\cite{alex2020really}, we introduce the notion of \textbf{conditional trust densities}.  Let a trust density $F(Q_z)$ be the distribution of question-answer trust $Q_z(x,y)$ for all questions $x$ that are answered as a given answer $y$.  To understand the distribution of question-answer trust at a finer granularity, we further decompose the trust density to gain insight into actor behaviour for a given oracle answer scenario where the answer is: 1) correct (i.e., $y=z$), and incorrect (i.e., $y \neq z$).  We accomplish this by computing $F(y=z)F(Q_z|y=z)$, the conditional density of question-answer trust given the answer is correct and $F(y \neq z)F(Q_z|y \neq z)$, the conditional density of question-answer trust given the answer is incorrect.  Note that the sum of the two conditional density curves equals $F(Q_z)$.

\section{Results and Discussion}
\label{results}

\begin{figure*}[t]
\centering
	\includegraphics[width = 0.8\linewidth]{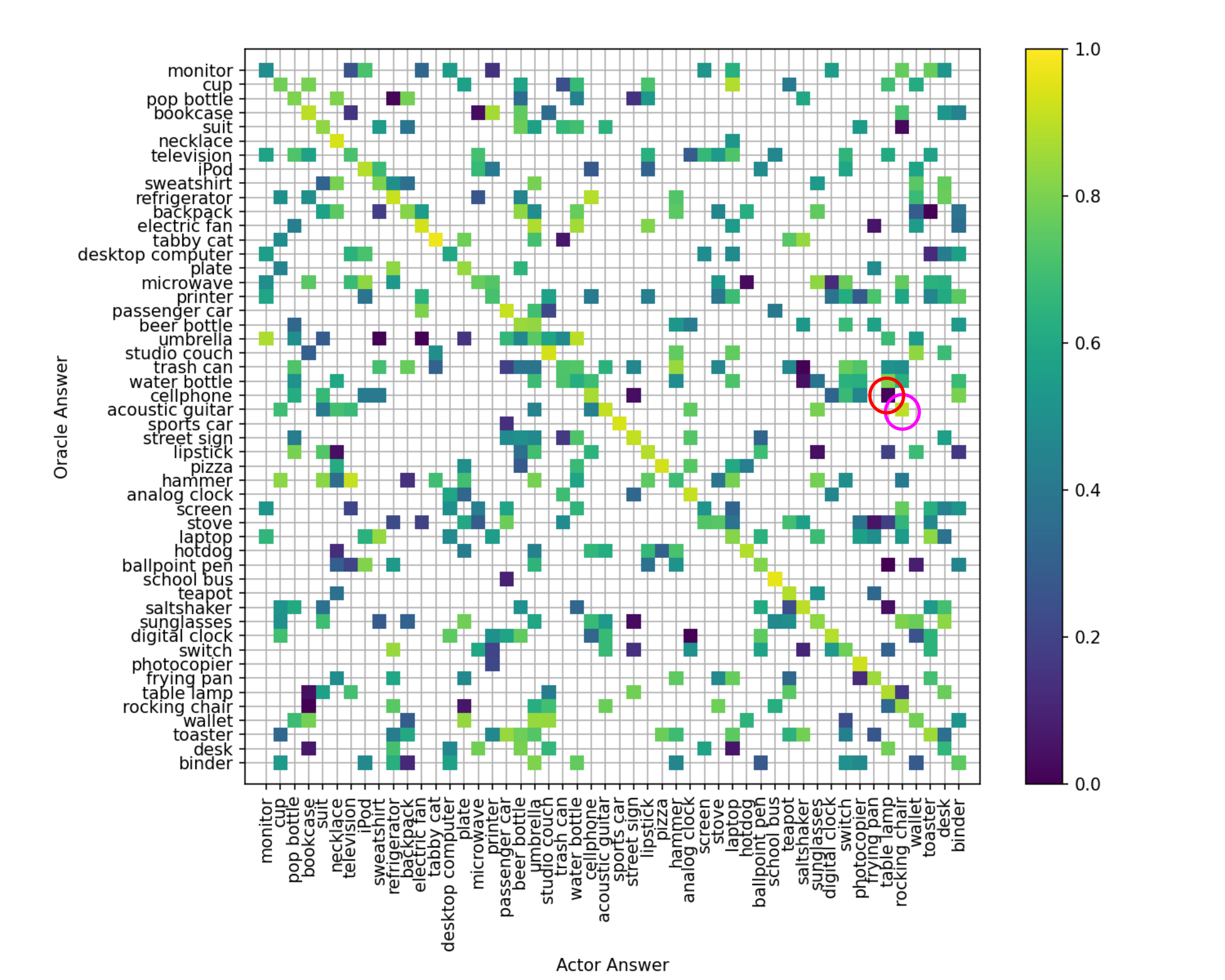}
	\caption{The trust matrix of MobileNet-V2 on a subset of the ImageNet~\cite{ImageNet} dataset. Red circle marks an overconfident actor-oracle answer scenario, while pink circle marks a scenario of strong trust despite being an incorrect answer scenario.}	
	\label{fig:matrix_mobilenetv2}
\vspace{-0.21 cm}
\end{figure*}

\begin{figure*}[t]
    \centering
    {\includegraphics[width=0.5\linewidth]{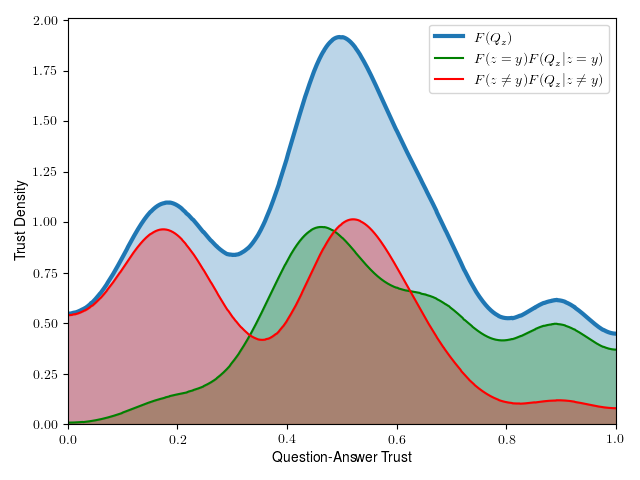}}
    \caption{Conditional trust densities of a ResNet-50 deep neural network for the specific oracle answer scenarios of `Monitor'.}
    \label{fig:resnet_monitor_trust_density}
\end{figure*}

To evaluate the efficacy of the proposed concept of trust matrix, we experimentally leverage trust matrices to study several well-known deep neural network architectures for image recognition, and further study the trust densities for an interesting actor-oracle answer scenario.  In this experiment scenario, the set of questions $x$ is a set of images the oracle asks the deep neural network to recognize, the answers $y$ are the predicted class labels, the oracle answers $z$ are the true class labels, and $C(y|x)$ for the answers $y$ are the softmax outputs related to the predicted class labels.  Similar to~\cite{alex2020really}, we leverage a subset of 2500 natural images from the ImageNet~\cite{ImageNet} benchmark dataset as a set of test questions to construct the trust matrices of the tested deep neural networks.  The well-known deep image recognition network architectures evaluated here are: 1) ResNet-50~\cite{ResNet50}, and 2) MobileNet-V2~\cite{MobileNetv2}.

\begin{table}[h]
	\centering
	\caption{NetTrustScore ($T_M$), conditional NetTrustScore for correct answers ($T_{M, y=z}$), and conditional NetTrustScore for incorrect answers ($T_{M, y \neq z}$) for two deep image recognition networks ResNet-50~\cite{ResNet50} and MobileNet-V2~\cite{MobileNetv2}.   }
	
	\begin{tabular}{p{3cm}|cccc|c}
		\hline
		Model ($M$)  & NetTrustScore ($T_M$) & $T_{M, y=z}$ & $T_{M, y \neq z}$\\
		\hline 
		ResNet-50~\cite{ResNet50} 	& 	0.776 	&	0.887 & 0.435 \\
		MobileNet-V2~\cite{MobileNetv2}	&  	0.739 & 0.845 & 0.507 \\	
		\hline
	\end{tabular}\\	
	\label{tab:Results}
\end{table}	

Table~\ref{tab:Results} shows the NetTrustScore for each of the models, as well as the corresponding conditional NetTrustScore for both correct and incorrect model answer scenarios. It can be observed that ResNet-50 outperforms MobileNet-V2 from an overall trustworthiness perspective.  However, looking at the conditional NetTrustScores, one can see that MobileNet-V2 achieves noticeably higher overall trustworthiness than ResNet-50 for incorrect answer scenarios, while ResNet-50 achieves noticeably higher overall trustworthiness than MobileNet-V2 for correct answer scenarios. The complexity of this trust trade-off illustrates the need for trust matrices (shown in Figures~\ref{fig:matrix_resnet50} and~\ref{fig:matrix_mobilenetv2} for ResNet-50 and MobileNet-V2, respectively) as a mechanism to further explore characteristics of a model's trustworthiness at a more fine-grained level.

A number of interesting observations can be made about the trustworthiness of the tested deep neural networks and where trust breaks down based on the trust matrices. First, the largely overconfident actor-oracle answer scenarios become very apparent upon visual inspection as low trust areas in the off-diagonal regions of the trust matrices (shown here as darker color regions). These areas of low trust along the off-diagonal regions of the trust matrices are prime targets for additional data collection as well as deep analysis. An example of such a overconfident actor-oracle answer scenario is `table lamp - cellphone' for MobileNet-V2 and `street sign - switch' for ResNet-50 (marked by a red circle).

Second, the strongly trusted scenarios (shown here as brighter color regions) are also made obvious upon visual inspection in the trust matrices, both along the diagonal (i.e., correct answer scenarios) and in the off-diagonal regions (i.e., incorrect answer scenarios).  In the scenarios of high trustworthiness but in the non-diagonal regions, a supplementary system (such as secondary classifier) could be built to help remedy the confusion while maintaining a high level of trust.  An example of such an actor-oracle answer scenario is `rocking chair - acoustic guitar' for MobileNet-V2 and `laptop - water bottle' for ResNet-50 (marked by a pink circle).

Third, it can be observed from the trust matrices that the two tested deep neural networks exhibit distinct trust behaviour with respect to the magnitude of the expected trustworthiness of any given oracle-actor answer tuple. More specifically, it can be observed that despite MobileNet-V2 having more incorrect answers and lower overall trust than ResNet-50, it exhibits noticeably larger proportions of high trust areas in the off-diagonal regions of the trust matrix than that of ResNet-50.  This is consistent with the higher conditional NetTrustScore for incorrect answer scenarios, and is a desirable trait to have from the perspective of identifying when to trust a prediction since a model's incorrect predictions are more easily detectable for intervention by another system (e.g., human).

We will now study the trust densities and conditional trust densities for several interesting actor-oracle answer scenarios identified using the trust matrices. Looking at the trust matrix of ResNet-50, one can see a large variation in expected scenario trusts along the diagonal. In the top left one can see the lowest expected trust in the `monitor' oracle answer scenario. Looking along the actor answers for the `monitor' oracle answer, a large collection of low expected trust scenarios can be seen. For the `monitor' oracle answer scenario, ResNet-50 is overcautious in its correct answers and overconfident in its incorrect answers. To further investigate the `monitor' oracle answer scenario, we visualize the its trust density and conditional trust densities in Figure~\ref{fig:resnet_monitor_trust_density}. The diagonal element in the trust matrix for the 'monitor' oracle answer scenario is expanded into $F(z=y)F(Q_z|z=y)$, and the off-diagonal scenarios are combined and expanded into $F(z \neq y)F(Q_z|z \neq y)$.

Moving from left to right along $F(z=y)F(Q_z|z=y)$, the low density characteristic of low trust regions is expected as correct answers are less likely to have very low trust levels while still being correct. The peak of $F(z=y)F(Q_z|z=y)$ is focused around the mid trust regions of the density curve. If we continue to move right along $F(z=y)F(Q_z|z=y)$, it can be seen that the conditional trust density begins to taper off. Looking at $F(z \neq y)F(Q_z|z \neq y)$, one can observe a bimodal distribution, with one peak at the very low trust region, and one peak that largely overlaps with the peak portion of $F(z=y)F(Q_z|z=y)$. The first mode of the distribution indicates large  overconfidence in incorrect answers which has a negative effect on the trustworthiness of ResNet-50 in being able to handle the 'monitor' oracle answer scenario. The large overlap between the peaks of each of the conditional trust densities is not a desirable characteristic to have as it makes distinguishing between correct and incorrect answers difficult. Overall, the trust density of ResNet-50 for the 'monitor' oracle answer scenario shows that it is not very trustworthy for this partricular scenario, as demonstrated by the low density in the high trust regions of the curve.

\section{Conclusions}
In this study, we introduced the concept of trust matrix for assessing the trustworthiness of deep neural networks by providing detailed insights of where trust breaks down at the actor-oracle answer scenario.  This provides an effective mechanism for spotting areas of low trust that need to be addressed.  We further introduced the concept of conditional trust densities for additional insights beyond standard trust densities at the oracle answer level. The proposed trust matrix and conditional trust densities were leveraged to study several well-known, complex deep neural network architectures for matrix the common task of image recognition using a subset of the ImageNet dataset, and led to interesting insights into where trust breaks down for the tested deep neural networks.  The results illustrate that trust matrices and conditional trust densities can be useful tools in addition to the existing suite of trust quantification metrics for guiding practitioners and regulators in creating and certifying deep learning solutions for trusted operation.

\bibliographystyle{IEEEtran}
\bibliography{references}
\end{document}